\documentclass[letterpaper, 10 pt, conference]{ieeeconf}  

\IEEEoverridecommandlockouts                              

\usepackage{graphicx}
\usepackage{caption}
\usepackage{subcaption}
\usepackage{amsmath} 

\title{\LARGE \bf
An Experience-based TAMP Framework for Foliated Manifolds
}

\author{Jiaming Hu$^*$ \and Shrutheesh R. Iyer$^*$ \and Henrik I. Christensen
\thanks{Contextual Robotics Institute, UC San Diego, La Jolla, CA 92093, USA}}

\begin{document}
\maketitle
\thispagestyle{empty}
\pagestyle{empty}

\begin{abstract}



Due to their complexity, foliated structure problems often pose intricate challenges to task and motion planning in robotics manipulation. To counter this, our study presents the ``Foliated Repetition Roadmap.'' This roadmap assists task and motion planners by transforming the complex foliated structure problem into a more accessible graph format. By leveraging query experiences from different foliated manifolds, our framework can dynamically and efficiently update this graph. The refined graph can generate distribution sets, optimizing motion planning performance in foliated structure problems. In our paper, we lay down the theoretical groundwork and illustrate its practical applications through real-world examples.

\end{abstract}

\section{INTRODUCTION}
Many robotics manipulation tasks can be split into multiple sub-tasks with unique constraints. For example, opening a door involves approaching and rotating its handle, each with distinct constraints. Such planning problems are called multi-modal problems, having several constraint manifolds, each modeling a sub-task. Many studies~\cite{englert2021rss, Fernandez2020LearningMF, hauser2011randomized} address these by planning motion trajectories across constraint manifolds. This falls under task and motion planning (TAMP), where the task planner first produces a task sequence at a high level, and then the motion planner solves each task geometrically.

\begin{figure}[t]
    \centering
    \includegraphics[width=0.45\textwidth]{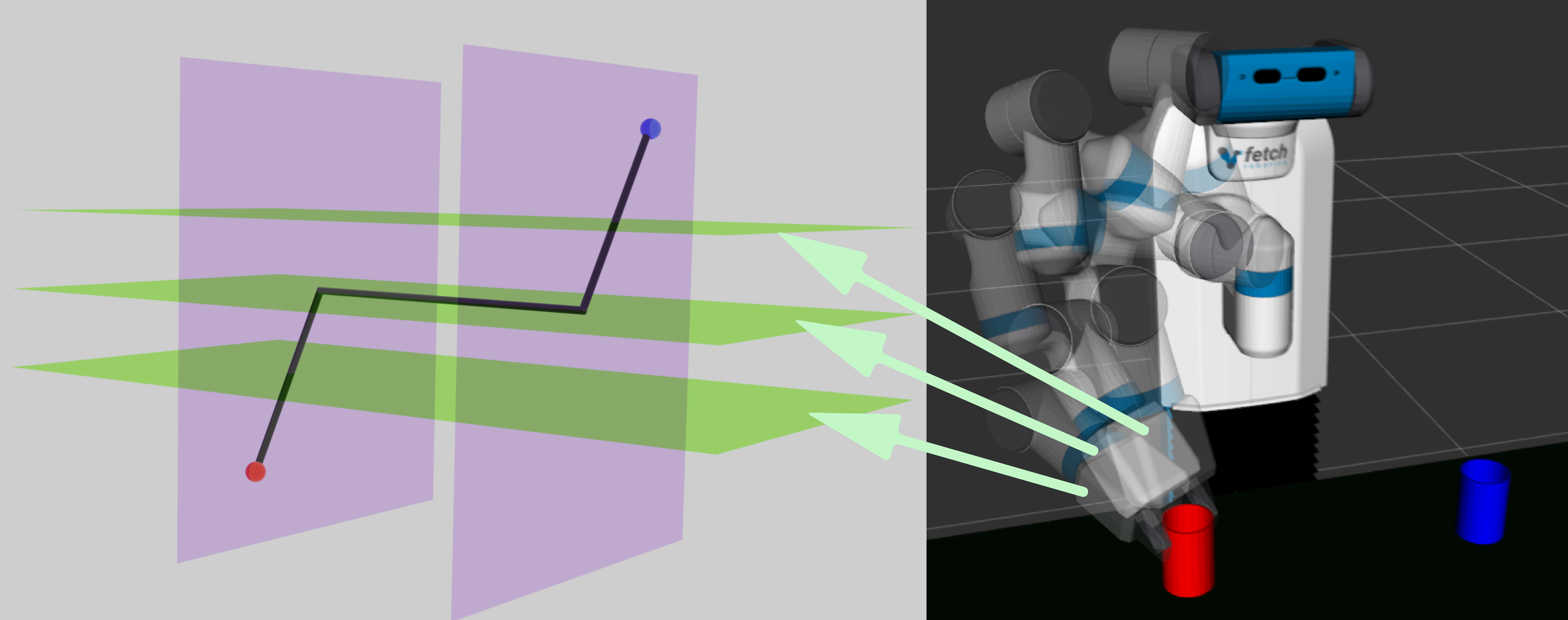}
    \caption{
   Delivering a cup horizontally (Right) can be considered as a foliated structure problem (Left). The initial object placement, represented by a red cup, and the target, symbolized by a blue cup, define the foliated manifolds—visualized as purple planes—during the ungrasping phases. Each potential grasp introduces another foliated manifold, depicted as green planes. The foliated structure problem is finding a path, the black line, that moves seamlessly across these foliated manifolds from the starting point (red dot) to the endpoint (blue dot).
    }
    \label{fig:simple_example}
\end{figure}

Many previous multi-modal planning works have opted not to consider the foliated structure (or foliated configuration space)~\cite{kim2014foliated} due to its complexity. This is a significant aspect of manipulation problems, where each sub-task should be defined by a set of manifolds, parameterized by continuous variables, such as different grasp poses or object placements.
For example, as shown in Fig.~\ref{fig:simple_example}, in delivering a cup of water, the cup must move horizontally, and the constraint on the robotic arm changes based on the specific grasp pose on the cup. We should employ foliation to address this intricate issue, where each grasp pose acts as a co-parameter to define a unique manifold.

The Mode Transition Graph (MTG)~\cite{kingston2020mtg} and Markov Decision Process (MDP)~\cite{hu23iros} are commonly used in the TAMP for foliated structure problems. New task sequences are generated by analyzing the status of past queries. By doing so, they hope to create sequences different from those previously failed, reducing the likelihood of failure in the next query. However, once a query fails, all planning experiences, such as sampled configurations with their status from the motion planner, are disregarded. This means the motion planner doesn't build a cumulative understanding of the problem with each successive query. 

To overcome this challenge, repetition roadmap~\cite{lehner2018roadmap} leverages experience from previous similar queries to plan more efficiently. However, it is designed specifically for a single manifold. Later,~\cite{kingston2021experience} points out that manifolds within a single foliation tend to be similar to each other if their co-parameters are ``nearby.'' This similarity allows the solutions from multiple manifolds to be applied as experiential knowledge when planning in a new similar manifold, potentially improving performance. However, it requires a solution to be present within similar manifolds. This prerequisite is not consistently met, which restricts its widespread use.

Thus, we present the ``Foliated Repetition Roadmap'' (FoliatedRepMap), an experience-driven TAMP framework that interprets planning experience and efficiently guides the motion planner to solve foliated structure problems. In essence, the FoliatedRepMap is an abstract roadmap seamlessly connecting the probability distributions to model the foliated manifolds of the problem. During the planning process, this roadmap can be refined using past motion planning experiences from similar foliated manifolds. This allows it to generate a probability distribution sequence that, when paired with the task sequence, guides the motion planner more efficiently to plan in new foliated manifolds.


In the upcoming sections, we first address related works in Sec.\ref{sec:related_work}. We provide essential preliminaries in Sec.\ref{sec:preliminaries}. The core of our methodology, which focuses on the FoliatedRepMap for MTG and MDP task planners in the context of the foliated structure problem, is discussed in Sec.\ref{sec:framework}. Finally, Sec.\ref{sec:experiments} showcases how our approach improves performance in different foliated structure problems.

\section{RELATED WORK} \label{sec:related_work}

\subsection{Multi-modal Planning}

Constraint motion planning generates paths that meet specific constraints. While techniques such as trajectory optimization-based methods~\cite{schulman2014motion} and sampled-based approaches~\cite{Berenson2009cbirrt} address planning within a single manifold constraint, navigating multiple constraint manifolds amplifies the challenge. This complexity stems from sequencing actions and corresponding valid motions. Some works, like~\cite{hauser2011randomized, hauser2010multi}, tackle multi-modal planning, but the task grows more arduous as the number of manifolds increases.

\subsection{Planning in Foliation}

As emphasized in~\cite{kingston2020mtg}, manifolds within a foliation with similar parameters often resemble each other. Utilizing this property, if a motion planner fails within a manifold $c$, the task planner can prioritize sequences excluding manifolds similar to $c$ to optimize the planning process in foliation.

Kingston presents the Mode Transition Graph~\cite{kingston2020mtg} (MTG) to model foliation problems graphically. Nodes depict manifold intersections, and edges symbolize path-planning tasks between these intersections under one manifold, each weighted by their planning difficulty. Using algorithms like Dijkstra's, the MTG task planner discerns optimal task sequences. When motion planning for a task fails, edge weights of similar tasks increase significantly; however, success leads to a slight edge weight increment. This design fosters diverse solution exploration and minimizes recurrent failures.

In our study presented in~\cite{hu23iros}, we put forth an alternative approach using the Markov Decision Process (MDP). Contrasting MTG's use of weights to indicate task difficulty, our MDP model employs probabilities to offer a more dynamic representation of the task's likelihood of success. With this probabilistic framework, we leverage value iteration to pinpoint the optimal task sequence. If a task fails during motion planning, associated probabilities for similar tasks are diminished. Conversely, successful planning amplifies these probabilities, setting the current edge's probability to 1.0. This methodology nudges the task planner towards previously successful strategies, facilitating an intelligent and adaptive reuse of effective solutions.

\subsection{Using Experience in Planning}

While numerous planners have been developed to tackle planning within foliation, the complexity of problems often leads to prohibitively long computation times. In response to this challenge, the paper~\cite{kingston2021experience} introduces a novel experience-based framework called ALEF, designed to enhance planning efficiency by drawing on planning experiences in similar manifolds. Although this approach successfully reduces planning time by transmitting waypoints as experiential data to the motion planner, it operates on the assumption that a solution exists within similar manifolds. Unfortunately, this condition is not always met, rendering the application of the framework infeasible in certain scenarios.

\section{PRELIMINARIES} \label{sec:preliminaries}


\subsection{Constraint planning in single manifold}
Planning within a single manifold involves finding a collision-free path from a starting point to a target point or region, all while adhering to the specific constraints of that manifold. Typically, given a continuous space \textbf{C}, these constraints are represented by a function 
$F$, with a configuration $q\in \textbf{C}$ satisfying this function if and only if $F(q) = 0$. In this context, the manifold comprises the set of all configurations that satisfy its constraint function. As a result, planning within a single manifold necessitates that every configuration along the chosen path must comply with the constraint function particular to that manifold. For a more in-depth exploration of this concept, refer to~\cite{kingston2018sampling}.

\subsection{Foliated manifolds}

In robotic manipulation, a task is often defined as a mode (or manifold) family, with continuous parameters that govern how a particular action is executed. Each specific parameterization delineates a unique mode (or manifold) within the family. Take, for example, the task of grasping a bar at any point along its length. In this scenario, every potential grasp location represents a distinct parameter, thereby defining a unique mode (or manifold) within the family. This collection of modes(or manifolds) is referred to as foliated manifolds~\cite{spivak1999comprehensive}. Unlike a general manifold, a foliated manifold incorporates a co-parameter. Given a co-parameter $\theta$, a foliated manifold can be expressed as
\[M_{\theta} = \{q \in \textbf{C} | F(q) = \theta\}\]

In addition to the aforementioned aspects, it is essential to note a critical property within foliated manifolds: the intersection between two manifolds originating from the same foliation should not exist.

\subsection{Repetition Roadmap in Motion Planning}

The primary objective of the repetition roadmap~\cite{lehner2018roadmap, lehner2017repetition} is to capture essential insights from prior solution paths within a roadmap under a single manifold. This knowledge is then applied to guide the search for new queries related to similar tasks. By utilizing a collection of experienced trajectories, an approximated Gaussian Mixture Model (GMM) is created, encompassing all relevant configurations. Subsequently, a graph is constructed, where nodes represent Gaussian distributions. Nodes are connected if there are sufficient trajectories passing through them, revealing relationships among the distributions. During planning, the start and goal distributions are identified using initial and target configurations. A sequence of distributions is generated through a shortest-path planning algorithm, offering foundational guidance to sampling-based motion planners, thereby significantly enhancing their efficiency.

\section{FRAMEWORK} \label{sec:framework}

\begin{figure} [t]
     \centering
     \begin{subfigure}[b]{0.22\textwidth}
         \centering
         \frame{\includegraphics[width=\textwidth]{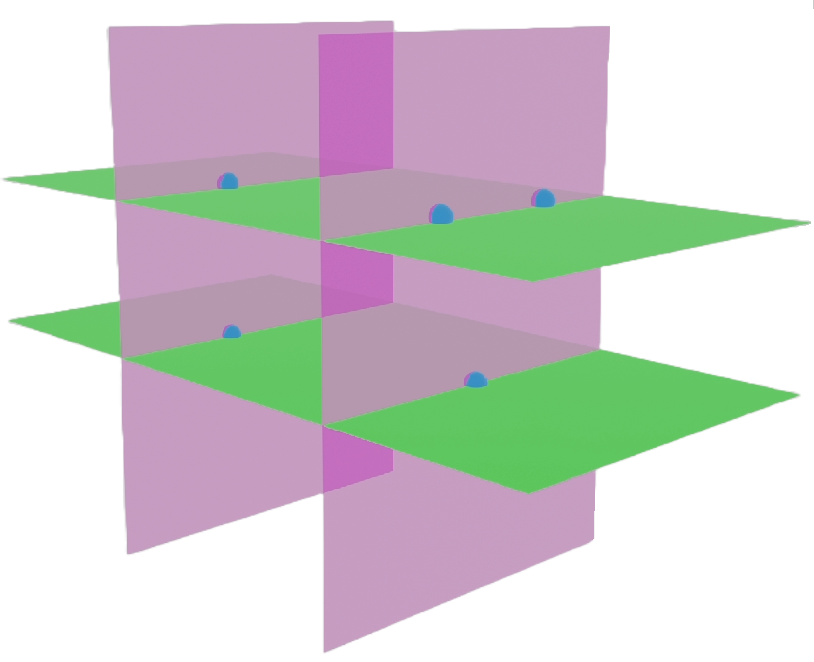}}
     \end{subfigure}
     \hfill
     \begin{subfigure}[b]{0.25\textwidth}
         \centering
         \frame{\includegraphics[width=\textwidth]{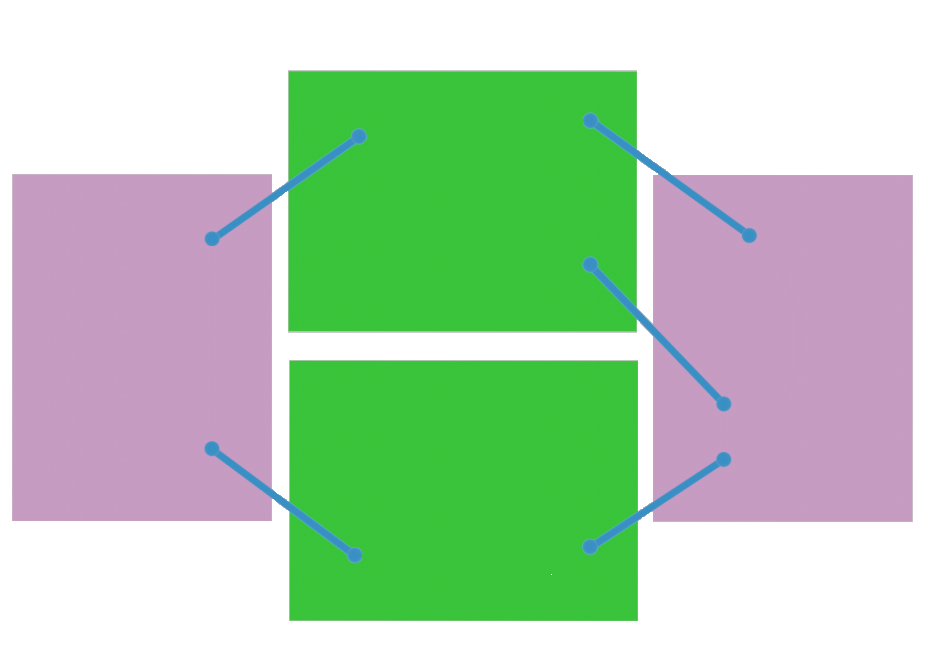}}
     \end{subfigure}
     \hfill
        \caption{Foliation example in different perspectives. Each plane is a manifold. The plane groups with the same color are a foliation, and here each foliation has two manifolds. The blue dots to the left (lines on the right) are the intersection between manifolds from different foliations.}
        \label{fig:foliation}
\end{figure}

In this section, we first provide a succinct introduction to our framework. Then, We outline the technique behind constructing the FoliatedRepMap, derived from the problem's foliated manifolds. Following that, we delve into the integration of FoliatedRepMap with both MTG and MDP task planners. Concluding the section, we will highlight the procedure used to update the FoliatedRepMap, incorporating feedback from the motion planner at each step.

\begin{figure}[t]
    \centering
    \includegraphics[width=0.3\textwidth]{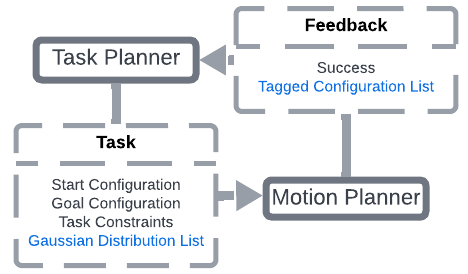}
    \caption{The main loop of our pipeline. Compared to earlier TAMP planners, our system integrates additional information, highlighted in blue, to enhance its performance.}
    \label{fig:main_loop}
\end{figure} 

\subsection{Main Pipeline}
Similar to most preceding TAMP frameworks, our framework also incorporates a two-level structure, consisting of a task planner and a motion planner. Given a foliation of the problem with a set of discovered intersections crossing between foliated manifolds, as shown in Fig.~\ref{fig:foliation}, the higher-level task planner generates a sequence of tasks to accomplish the desired manipulation, where each task contains a start configuration and a target configuration with the task constraints. As shown in Fig.~\ref{fig:main_loop}, what sets our approach apart from previous works is the inclusion of a Gaussian distribution list for each task. This set functions as a region where the motion planner is more likely to sample, due to a higher prior, thus providing an enhancement in the motion planning process.

On the other hand, the lower-level motion planner accepts the task and plans a trajectory using the constrained Bi-directional RRT algorithm\cite{berenson2009cbirrt}. Rather than merely sampling uniformly, our planner allocates a configuration based on the Gaussian distribution list with a probability of $\rho$, and complements this with uniform sampling at a rate of $1 - \rho$. As part of an experience-based framework, our motion planner must also return sets of both invalid and valid configurations as planning experience. Specifically, during our configuration evaluation step, the motion planner must identify the reason for an invalid sampling. In this project, we only consider the following invalid sampling reasons:

\begin{enumerate}
  \item Collision between robot and environment.
  \item Collision between the manipulated object and the robot or the environment.
  \item Violation of task constraints.
\end{enumerate}

Upon completing the motion planning, the motion planner will return the feedback including a success flag and a list of configurations, each accompanied by a tag indicating whether the configuration is valid or not, and if invalid, the specific reason for its invalidity. 

\begin{figure} [t]
     \centering
    \includegraphics[width=0.46\textwidth]{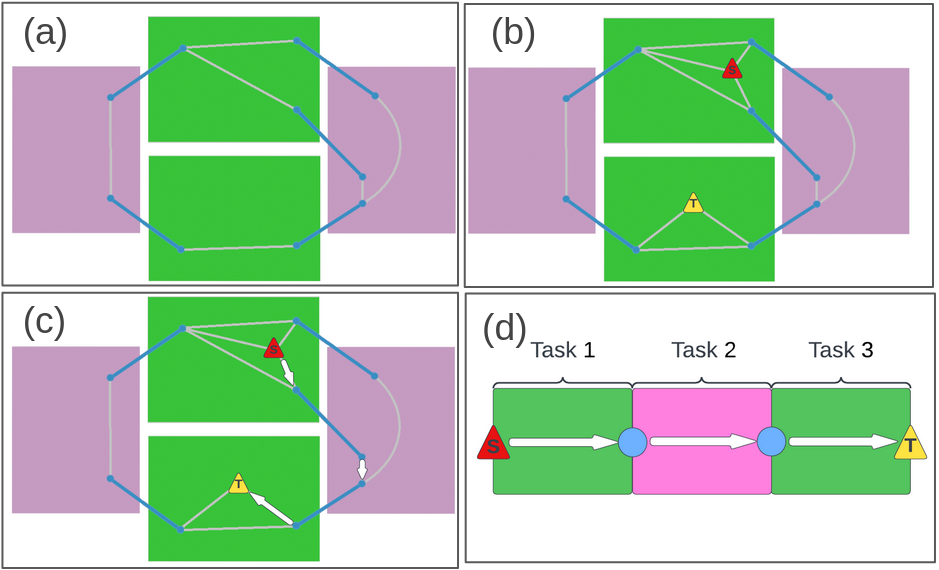}
    \caption{Task sequence generation of previous works. (a) Converting a foliated structure problem into a graph with intersections as nodes and possible motion between two intersections as edges. (b) Given the start (red triangle) and target (yellow triangle) configurations in different manifolds, add them to the graph. (c) Search the optimal node sequence from start to target. (d) Convert the optimal node sequence to the task sequence.}
    \label{fig:task_graph}
\end{figure}

\subsection{Constructing FoliatedRepMap}

Researchers often address the complexities of the foliated structure problem by transforming it into a more tractable graph problem. In previous research~\cite{kingston2020mtg, hu23iros}, scholars constructed the foliated manifolds into a graph within their task planner, as shown in Fig.~\ref{fig:task_graph}. Their method treats intersections between two different manifolds as nodes, and edges connect two nodes from different manifold pairs. During planning, the start and goal configurations are nodes connected to all intersections of their manifolds. Thus, by employing various optimal path planning algorithms, a node sequence is produced, and each edge in this sequence is thought of as a task, with its two end nodes representing the starting and goal configurations. The constraints of the manifold where the edge lies are taken as the task constraints. Thus, the task planners generate a task sequence detailing which manifolds to cross. 

In our framework, we construct the graph differently. Similar to~\cite{lehner2018roadmap, lehner2017repetition}, building a repetition roadmap in one single manifold, we construct a new repetition roadmap, named ``foliated repetition roadmap,'' across multiple manifolds.

\begin{figure}[t]
    \centering
    \includegraphics[width=0.4\textwidth]{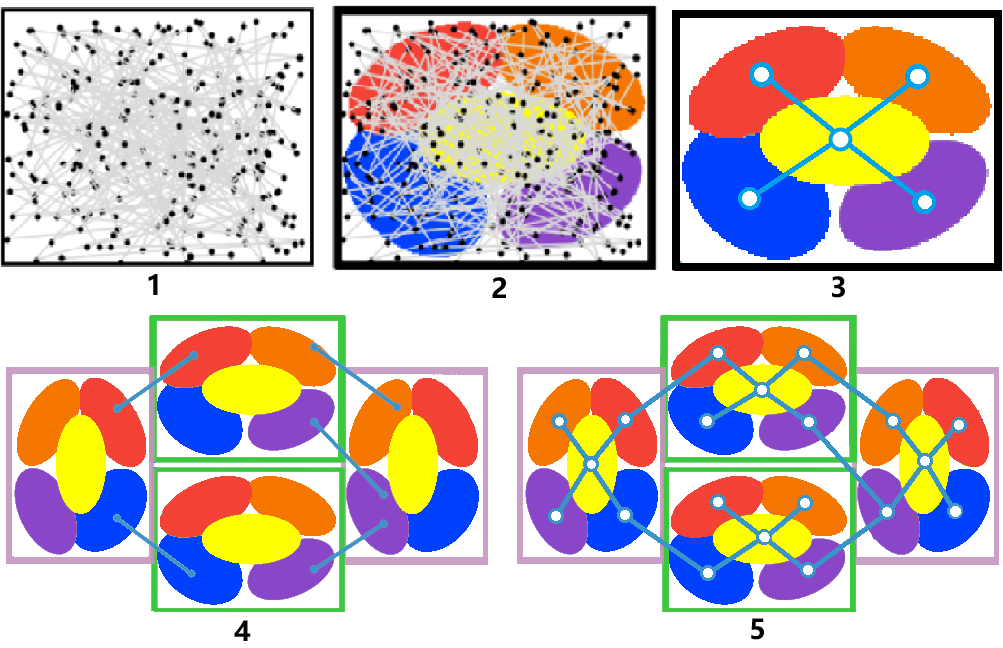}
    \caption{(1) Collect trajectory dataset from self-collision-free space. (2) Estimate GMM (each ellipse is a Gaussian distribution) to model self-collision-free space. (3) Build a FoliatedRepMap for each foliated manifold. (4) Get the intersections (blue lines) between foliated manifolds. (5) Connect all FoliatedRepMaps based on those intersections.}
    \label{fig:foliated_repetition_roadmap}
\end{figure} 

\subsubsection{Building FoliatedRepMap in Single Manifold}
First, we build the FoliatedRepMap for the self-collision-free space in one manifold without constraints, similar to~\cite{johnson2023learning}. As shown in (1) of Fig.~\ref{fig:foliated_repetition_roadmap}, we start by collecting a dataset comprising self-collision-free trajectories generated by running RRT star~\cite{karaman2011sampling} on pairs of random valid configurations. We then apply a clustering algorithm, such as the Dirichlet process algorithm (based on~\cite{huh2019thesis}, opting for this over the Expectation-Maximization algorithm used in~\cite{lehner2018roadmap}), to all the waypoints of the dataset. This produces ((2) of Fig.~\ref{fig:foliated_repetition_roadmap}) a Gaussian Mixture Model (GMM) that approximates the robot's self-collision-free configuration space.

However, at this stage, no edges exist between the distributions. Therefore, for each trajectory within the dataset, we identify the sequence of distributions $\{d_0, ..., d_{i}, d_{i+1}, ...\}$ that the trajectory traverses. For every pair of adjacent distributions $d_{i}$ and $d_{i+1}$ along the trajectory, the trajectory indicates a connection between these two distributions. If a sufficient number of trajectories suggest that two distributions are linked, an edge is created to connect them, as shown in (3) of Fig.~\ref{fig:foliated_repetition_roadmap}. Eventually, a FoliatedRepMap exists in each foliated manifold.

\subsubsection{Connecting FoliatedRepMaps between Manifolds}

Every manifold contains a identical FoliatedRepMap. To represent this structure, each FoliatedRepMap node is defined as $n_{i, \theta, j}$ where $i$ is the foliation, $\theta$ is the co-parameter associated with foliation $i$, and $j$ is the distribution. Together, $i$ and $\theta$ specify a foliated manifold $M_{i, \theta}$. 

At the current stage, the FoliatedRepMaps within different manifolds remain isolated. To establish connections between them, we need to utilize the intersections between the manifolds (Blue lines in (4) of Fig.~\ref{fig:foliated_repetition_roadmap}). Consider two foliated manifolds $M_{a,b}$ and $M_{c,d}$ from distinct foliations, along with their intersections. For each intersection, we identify the distribution $j$ in which it resides. This allows us to create a connection between $n_{a, b, j}$ and $n_{c, d, j}$ based on the given intersection. By applying this method to all relevant intersections, we can unify the separate FoliatedRepMaps, weaving them into a coherent, interconnected structure, as shown in (5) of Fig.~\ref{fig:foliated_repetition_roadmap}.

\subsection{Planning with FoliatedRepMap}

Upon constructing the FoliatedRepMap, the task planner must efficiently harness this roadmap to generate task sequences. In this section, given a starting configuration $q_{start}$ within the foliated manifold $M_{s, s'}$ and a goal configuration $q_{goal}$ within $M_{g, g'}$
 , we'll showcase how two distinct task planners—the MTG and MDP—derive the optimal task sequence from the FoliatedRepMap.

\subsubsection{MTG Task Planner}

\begin{figure}[t]
    \centering
    \includegraphics[width=0.4\textwidth]{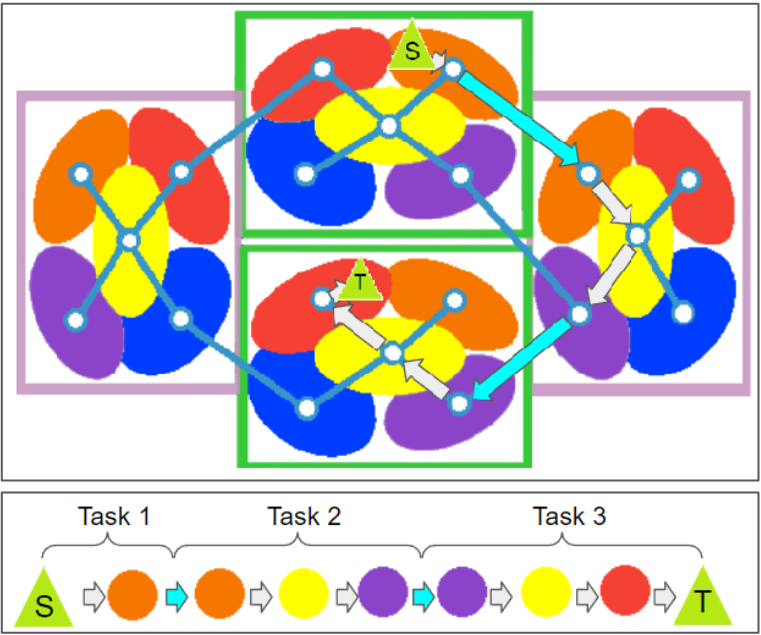}
    \caption{Generate task sequence with FoliatedRepMap. Upper: Add the start and goal configurations in different manifolds into the roadmap and search for a node sequence. Lower: Given the node sequence, split them into a task sequence by the intersection edge (shown as light blue arrows). Colored circles under the task are probability distributions used in motion planning for that task.}
    \label{fig:plan_foliated_repetition_roadmap}
\end{figure}

In the MTG task planner, each edge of the FoliatedRepMap is assigned a weight that indicates the difficulty of transitioning between the distributions of two nodes. A higher weight signifies greater difficulty. To generate an optimal task sequence, we first identify the distributions $d_{start}$ and $d_{goal}$ to which $q_{start}$ and $q_{goal}$ belong, respectively. Subsequently, we designate $n_{s, s', d_{start}}$ as the starting node and $n_{g, g', d_{goal}}$ as the goal node. The task planner then employs the Dijkstra's algorithm, using the edge weights, to determine the shortest path, resulting in a sequence of nodes $\{n_1, n_2, ...\}$, as shown in Fig.~\ref{fig:plan_foliated_repetition_roadmap}.

Given that the node sequence might span multiple manifolds, intersections in the sequence segments it into distinct sections. Each of these sections comprises nodes from a single manifold. Every section represents a task, with the intersections on its boundaries serving as the starting and ending configurations. The node distributions within a section are then relayed to the motion planner as prior information for sampled-based motion planning.

\subsubsection{MDP Task Planner}
In the MDP task planner, each edge of the FoliatedRepMap carries a probability indicating the likelihood that the motion planner can identify a solution path between the distributions of two nodes. To determine the optimal task sequence, much like the approach in the MTG task planner, we first identify both the starting and goal nodes. Adhering to the principles of MDP, we treat each node in the FoliatedRepMap as a state, and each edge as an action. For an agent positioned in state $s$, selecting an action $a$ leading to state $s'$ could have a certain probability of reaching a dead-end state, resulting in a penalty. Conversely, reaching the goal node yields a positive reward. With this MDP graph, we utilize value iteration to determine an optimal node sequence. This sequence is then transformed into a task sequence, using the same method as the MTG task planner.

\subsection{Updating FoliatedRepMap}
After the motion planning phrase on a task, our motion planner returns a set of tagged sampled configurations as feedback, so the task planner can process this feedback to adjust the FoliatedRepMap for subsequent replanning.

To receive the feedback, in the FoliatedRepMap, each node $n$ contains the following counts:
\begin{enumerate}
  \item $C_{valid}^n$: Count of valid arm configuration.
  \item $C_{robot-invalid}^d$: Count of invalid arm configuration because of the collision between arm and environment, and $d$ is the distribution. A key point to note is that all nodes with the same distribution collectively share a single $C_{robot-invalid}^d$ value.
  \item $C_{object-invalid}^n$: Count of invalid arm configuration because of the collision between object and either arm or environment.
  \item $C_{const-invalid}^n$: Count of invalid arm configuration because of constraint violation.
\end{enumerate}
After motion planning, those counts will be increased respectively based on the feedback. Consider a task planning within the foliated manifold $M_{i, \theta}$. The motion planner would produce a series of tagged arm configurations, represented as $\{(q_1, t_1), (q_2, t_2), ...\}$. Each configuration $q$ allows us to identify the distribution $j$ to which $q$ belongs. The tag $t$ indicates the status of $q$. Accordingly, we increment the relevant count of node $n_{i, \theta, j}$. Specifically:

\begin{enumerate}
  \item $q$ is valid, increment $C_{valid}^{n_{i, \theta, j}}$ by one.
  \item $q$ is invalid due to either collision between the robot and the environment, increment $C_{robot-invalid}^j$ by one.
  \item $q$ is invalid due to the collision on the manipulated object, increment $C_{object-invalid}^{n_{i, \theta, j}}$ by one.
  \item  $q$ is invalid due to constraint violation, increment $C_{const-invalid}^{n_{i, \theta, j}}$ by one.
\end{enumerate}

After updating the node counts, the task planners refine the edge values within the FoliatedRepMap for subsequent re-planning. Remembering that updates should not be confined to a single manifold is crucial. Instead, values across various manifolds must also be adjusted, reflecting the inherent properties of foliation.

\subsubsection{MTG Task Planner}

Given a node $n_{i, \theta, j}$, it has an associated score $s_{n_{i, \theta, j}}$ which is calculated in the following way:
\[s_{n_{i, \theta, j}} = C_{robot-invalid}^j * v^+ +\]
\[\sum_{\theta'}S_{\theta, \theta'}[v^-C_{valid}^{n_{i, \theta', j}} + v^+(C_{object-invalid}^{n_{i, \theta', j}} + C_{const-invalid}^{n_{i, \theta', j}})]\]
where $v^-$ is a small penalty, $v^+$ is a large penalty, $\theta'$ is all co-parameters in foliation i, and $S_{\theta, \theta'}$ is the similarity value(ranges from 0 to 1, with higher values indicating greater similarity.) between $\theta$ and $\theta'$. Ultimately, the weight of an edge between nodes m and n should be:
\[weight = s_m + s_n\]

\subsubsection{MDP Task Planner}
Given a node $n_{i, \theta, j}$, it contains a valid score $s^+_{n_{i, \theta, j}}$ and a invalid score $s^-_{n_{i, \theta, j}}$ where 

\[s^+_{n_{i, \theta, j}} = \sum_{\theta'}S_{\theta, \theta'}C_{valid}^{n_{i, \theta', j}}\]
\[s^-_{n_{i, \theta, j}} = C_{robot-invalid}^j + \]
\[\sum_{\theta'}S_{\theta, \theta'}(C_{object-invalid}^{n_{i, \theta', j}} + C_{const-invalid}^{n_{i, \theta', j}}) \]
Then, given an edge between two nodes m and n, its probability is defined as follows:
\[
    probability = 
    \begin{cases}
        0.5 \text{ if $s^+_m + s^+_n + s^-_m + s^-_n = 0$,}\\
        \frac{1 + s^+_m + s^+_n}{1 + s^+_m + s^+_n + s^-_m + s^-_n}\text{ otherwise.}
    \end{cases}
\]
Once the FoliatedRepMap is updated, both task planners are equipped to re-plan a new optimal task sequence.

\begin{figure}[t]
    \centering
    \includegraphics[width=0.45\textwidth]{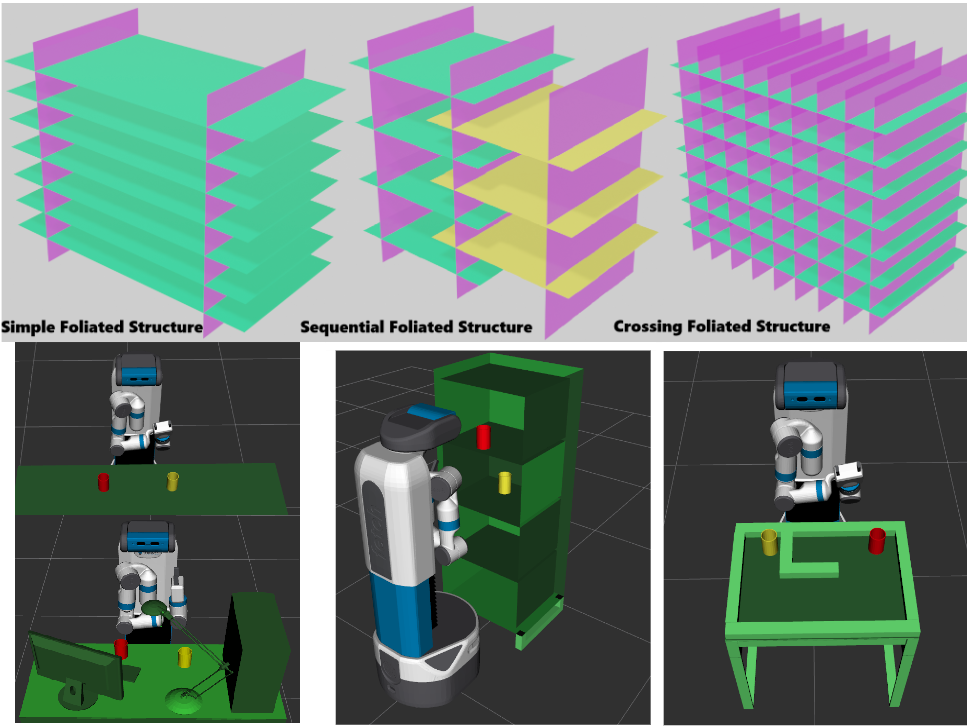}
    \caption{Different foliated structure problems. i) Simple foliated structure: Sliding a cup with or without obstacles, where grasps define green foliated manifolds and start/end placements define pink foliated manifolds. ii) Sequential foliated structure: Delivering a cup from the lower to the upper part of a shelf by dragging and re-grasping. The intermediate placement for re-grasping defines the central purple foliated manifold, while green and yellow foliated manifolds are defined by grasps for dragging and delivering. iii) Crossing foliated structure: Sliding a cup in a maze with re-grasping. Each intermediate placement for re-grasping defines a purple foliated manifold, while each grasp for sliding defines a green foliated manifold.}
    \label{fig:all_foliated_structure}
\end{figure} 


\section{EXPERIMENTS} \label{sec:experiments}
This section will set up the experiment to compare the task planner with FoliatedRepMap and without it on different foliated structure problems.

\subsection{Setup}

Most manipulation problems with foliated structure can be classified into three categories as shown in Fig.~\ref{fig:all_foliated_structure}: i) Simple foliated structure, ii) Sequential foliated structure, and iii) Crossing foliated structure.
In this section, we designed three different setups to represent each structure. 
For a simple foliated structure, the task is sliding a cup on a desk without re-grasping. For sequential foliated structure, the task is dragging a cup from the inside of the lower part of a shelf to the outside, then re-grasping and delivering it horizontally to the upper part of the shelf. For the crossing foliated structure, the task is sliding a cup from one location to another in a maze while keeping it in contact with the table. Here, we adopt a method from Wan~\cite{wan2015regrasp} to generate grasps (100 grasps in our experiments) for manipulation. Each of these grasps defines a foliated manifold. Additionally, pose differences is the metric to gauge the similarity between co-parameters.

For each of the setups, our experiment runs 50 times and records the success rate, running time, and total distance the arm has traveled. For each loop, the task planner has two seconds of planning time. If the system has done more than 100 loops, then we consider it a timeout failure.

\begin{table}
\caption{Running Time(s) with Different Initialization}
\label{table: DL-table}
\begin{center}
\begin{tabular}{||c c c c||} 
 \hline
 Task & Simple & Sequential & Crossing \\ [0.5ex] 
 \hline\hline
 Uniform Initialization & 91.28 & 100.044 & 77.009 \\ 
 \hline
 Initialization w/ Predictor & 48.54 & 77.107 & 62.984 \\
 \hline
\end{tabular}
\end{center}
\end{table}

\subsection{Planning Results in Different Foliated Structures}
\begin{figure}[t]
    \centering
    \includegraphics[width=0.4\textwidth]{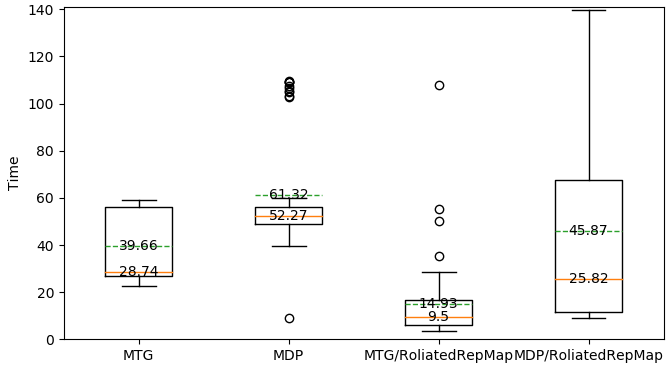}
    \caption{Planning time(s) in the relatively easy simple foliated structure without obstacles. (Orange lines: median values, Green lines: mean value).}
    \label{fig:time_dis_pick_and_place_with_constraint}
\end{figure}
For a fair comparison of the running times across different task planners, we must select a task where every planner achieves 100\% success rate. Accordingly, the simple foliated structure problem of sliding the cup on a desk without obstacles is chosen as the baseline. Analyzing Fig.~\ref{fig:time_dis_pick_and_place_with_constraint} makes it clear that within this task, the MTG task planners consistently outpace the MDP task planners. It's worth highlighting that when used with the FoliatedRepMap, the MTG task planner delivers the most superior performance.

\begin{figure}[t]
    \centering
    \includegraphics[width=0.45\textwidth]{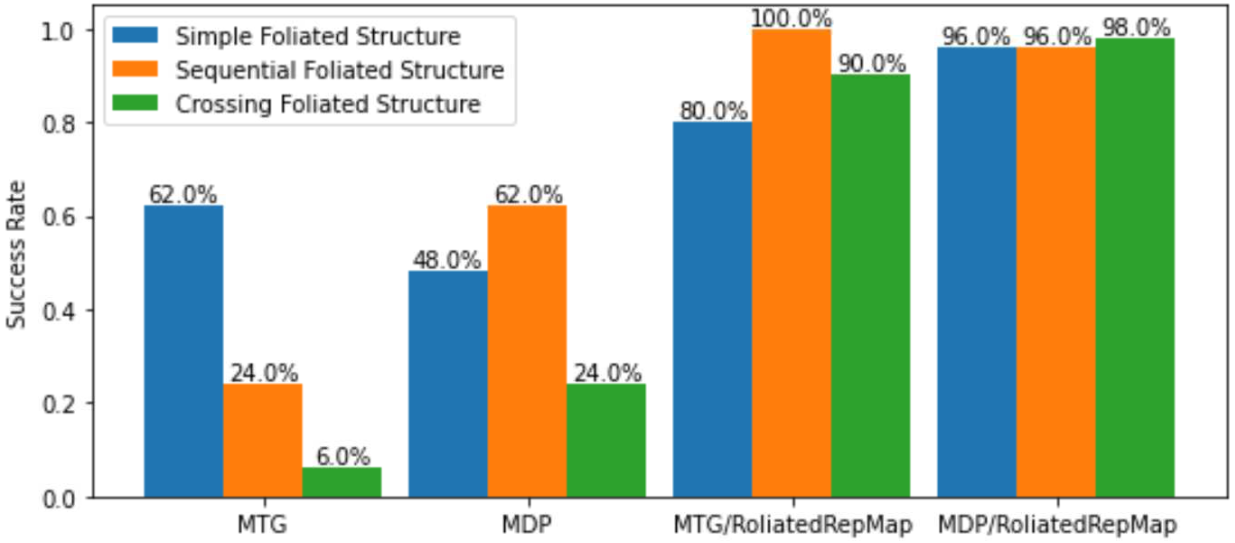}
    \caption{Success rates in different problems.}
    \label{fig:success_rates}
\end{figure}

For problems with increased complexity, the success rates of the planners are compared. As shown in Fig.~\ref{fig:success_rates},
as the complexity increases 
, there is a significant drop in success rates for task planners that lack the FoliatedRepMap. On the other hand, task planners integrated with FoliatedRepMap consistently outperform those without it, especially in sequential and crossing foliated structures.


\subsection{Other Observations}
In our experiments, there are additional observations. First, the total traveling distance doesn't have a significant improvement with FoliatedRepMap in all different problems. Second, in all kinds of foliated structure problems, task planners equipped with FoliatedRepMap exhibit similar success rates. However, notable differences emerged during the experiment. The MTG task planner is consistently faster than the MDP task planner, which can be attributed to its superior task sequence generation. Typically, Dijkstra's algorithm, employed by MTG, has a shorter runtime than the value iteration used by MDP. Third, the MTG task planner generally requires minimal parameter tuning.
In contrast, the MDP task planner demands extensive adjustments to its MDP parameters. This is due to the large state space, which prevents gradients from traveling across the MDP from the positive reward (target) state to the initial state, leading to non-convergence. Finally, in unconstrained (or with easy constraints) problems, like pick-and-place on a shelf without moving it horizontally, task planners equipped with FoliatedRepMap don't exhibit notable advantages over planners that lack it.

\subsection{Potential Improvement}
One potential avenue to improve the performance of the system is to provide a more informed initialization of the graph for faster convergence. This can be achieved by increasing the probability (or decreasing the weight) of distributions that are more likely to be collision-free. Using point cloud data, the collision scores of each distribution can be predicted (by learning with synthetic collision environments). Preliminary work in this direction has yielded promising results, as shown in Table~\ref{table: DL-table}, which are the running times, collected from a different computer, of MTG task planner with FoliatedRepMap with different initialization.

\section{CONCLUSIONS}
In this article, we introduce our framework, the ``Foliated Repetition Roadmap'', designed for planning problems involving foliated structures. We demonstrate the integration of this framework into multiple task planners for different tasks. Experimental results indicate that our framework building on MTG Task planner has the best success rate and speed in difficult foliated structure problems.

\newpage

\addtolength{\textheight}{-12cm}   




\end{document}